%% file: root.tex
\documentclass[letterpaper, 10 pt, conference]{ieeeconf}  

\IEEEoverridecommandlockouts                              

\usepackage{graphics} 
\usepackage{epsfig} 
\usepackage{mathptmx} 
\usepackage{times} 
\usepackage{amsmath} 
\usepackage{amssymb}  

\usepackage{subcaption}
\usepackage{hyperref}
\usepackage[capitalize, noabbrev]{cleveref}
\crefname{section}{Sec.}{Secs.}
\Crefname{section}{Section}{Sections}
\Crefname{table}{Table}{Tables}
\crefname{table}{Tab.}{Tabs.}
\crefformat{section}{\S#2#1#3}
\crefformat{subsection}{\S#2#1#3}
\crefformat{subsubsection}{\S#2#1#3}

\usepackage{xcolor}
\usepackage{amsmath,amsfonts,bm, bbm}
\usepackage{float}
\usepackage{multirow}
\usepackage{booktabs}
\usepackage{dsfont}

\title{\LARGE \vspace{-5pt} \bf
A Multi-Loss Strategy for Vehicle Trajectory Prediction: \\
Combining Off-Road, Diversity, and Directional Consistency Losses \vspace{-5pt}
}

\author{Ahmad Rahimi$^{1}$ and Alexandre Alahi$^{1}$
\thanks{*This work was funded by Hasler Foundation under the Responsible AI.}
\thanks{$^{1}$EPFL, Lausanne, Switzerland; Email: firstname.lastname@epfl.ch}
}

\newcommand{\norm}[1]{\ensuremath{\| #1 \|_2}}

\begin{document}

\maketitle
\thispagestyle{empty}
\pagestyle{empty}

\begin{abstract}
    \input{sections/abstract}
\end{abstract}

\section{INTRODUCTION}
\label{sec:intro}
\input{sections/intro_final}

\section{RELATED WORK}
\label{sec:related}

\input{sections/related}

\section{METHOD}
\label{sec:method}
\input{sections/method}

\section{EXPERIMENTS}
\label{sec:experiment}
\input{sections/experiment}

\section{Conclusions}
\label{sec:conclusion}
\input{sections/conclusion}


\newpage
\addtolength{\textheight}{-12cm}   
\bibliographystyle{IEEEtran}
\bibliography{bibtex/root}

\end{document}

%% file: sections/abstract.tex
Trajectory prediction is essential for the safety and efficiency of planning in autonomous vehicles. However, current models often fail to fully capture complex traffic rules and the complete range of potential vehicle movements. Addressing these limitations, this study introduces three novel loss functions: Offroad Loss, Direction Consistency Error, and Diversity Loss. These functions are designed to keep predicted paths within driving area boundaries, aligned with traffic directions, and cover a wider variety of plausible driving scenarios. As all prediction modes should adhere to road rules and conditions, this work overcomes the shortcomings of traditional ``winner takes all" training methods by applying the loss functions to all prediction modes. These loss functions not only improve model training but can also serve as metrics for evaluating the realism and diversity of trajectory predictions. Extensive validation on the nuScenes and Argoverse 2 datasets with leading baseline models demonstrates that our approach not only maintains accuracy but significantly improves safety and robustness, reducing offroad errors on average by 47\% on original and by 37\% on attacked scenes. This work sets a new benchmark for trajectory prediction in autonomous driving, offering substantial improvements in navigating complex environments. Our code is available at \url{https://github.com/vita-epfl/stay-on-track}.

%% file: sections/intro_final.tex
Trajectory prediction plays a crucial role in autonomous systems, particularly in enhancing the safety and reliability of self-driving vehicles. Recent advancements have significantly improved the capabilities of predictive models, enabling them to generate multimodal trajectory predictions that account for uncertainties in drivers' intentions and driving styles. Despite these advancements, significant challenges remain, including accurately understanding scenes and predicting all possible trajectory modes.

\begin{figure}[t!]
  \centering
  \includegraphics[width=0.85\linewidth]{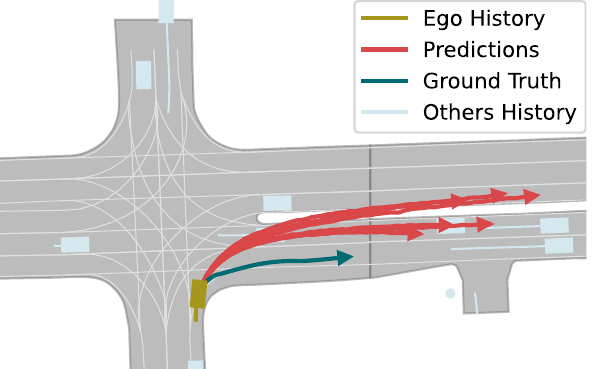}
  \caption{Trajectory predictions by Wayformer \cite{nayakanti_wayformer_2022}, a state-of-the-art model, highlighting errors such as off-road movements, incorrect traffic direction adherence, and missed predictions for other plausible maneuvers at the intersection, such as continuing straight. Our proposed loss functions aim to correct these prediction errors.}
   \label{fig:pull}
   \vspace{-10pt}
\end{figure}

Current models often struggle with fully grasping scene dynamics and accurately forecasting every potential movement, as evidenced in \cref{fig:pull}. Even advanced models may incorrectly predict trajectories that go off-road, oppose traffic flow, or overlook viable maneuvers at intersections and in complex urban settings. This highlights the need for enhanced models capable of better interpreting and navigating various traffic situations for improved safety and dependability.

Moreover, traditional objective functions used in model training often fall short in enhancing the mentioned aspects of prediction. Typically, these training methods adopt a ``winner takes all" approach where they focus on minimizing the prediction errors for the trajectory closest to the ground truth observation. This method neglects other viable trajectory predictions during the model's learning phase, as only the closest prediction receives updates from the gradients. This can result in suboptimal predictions of less frequent yet feasible trajectories, thus limiting the model's ability to handle rare or complex maneuvers effectively.

In this work, we address this gap by focusing on the quality of all generated trajectories. This study introduces three loss functions aimed at embedding more targeted inductive biases into trajectory prediction models. Our proposed losses — Offroad Loss, Direction Consistency Error, and Diversity Loss — operate across all prediction modes. They are designed to deepen the model's scene comprehension, ensure adherence to traffic regulations, and enhance the diversity of trajectory predictions.

Specifically, the \textit{Offroad Loss} ensures that trajectories stay within drivable boundaries, imposing increasing penalties for deviations further from these zones to encourage accurate navigation. The \textit{Direction Consistency Error} loss aligns predicted trajectories with the expected traffic flow and road layouts. It is particularly beneficial in complex environments such as intersections and multi-lane roads with opposing traffic flows, where it plays a crucial role in guiding the vehicle safely through the correct path. Finally, the \textit{Diversity Loss } encourages the model to generate a varied set of predictions that cover all plausible outcomes, crucial for dynamic traffic situations where multiple routes may be feasible. It selectively filters out infeasible predictions, and calculates the sum of pairwise distance between remaining predictions.

In summary, this work introduces three novel loss functions devised to address key shortcomings in recent trajectory prediction models. Each function specifically targets a distinct aspect of prediction quality, from adherence to drivable paths and alignment with traffic flow, to the diversity of outcomes. These functions are uniformly applied across all modes, overcoming the constraints of ``winner takes all" methods and ensuring comprehensive refinement of all potential trajectories during training, thereby improving the precision and safety of the model’s predictions.

We conducted extensive experiments using two leading baseline models on the nuScenes and Argoverse 2 datasets to assess the effectiveness of our proposed loss functions. The results confirm that these functions enhance the robustness and safety of the trajectory prediction models without sacrificing prediction accuracy. Notably, we have achieved a significant reduction in the offroad metric by half on original dataset scenarios, and 37\% when adding natural turns in the scene, underscoring the effectiveness of our approach.

%% file: sections/related.tex
Recent advancements in vehicle trajectory prediction have led to significant improvements in modeling techniques. Inspired by breakthroughs in convolutional neural networks (CNNs) and computer vision, early models utilized rasterized bird's-eye view images, incorporating static map elements and agent trajectories as input representations. This approach leveraged CNNs to predict future paths
\cite{ren2021safety, biktairov2020prank, chai2019multipath, hong2019rules, casas2018intentnet}. However, rasterized inputs are computationally intensive and memory-heavy, suffer from a limited field of view, and burden the model with extracting already available 2D positional information — challenges that complicate agent interaction modeling and scene comprehension.

Addressing these limitations, more recent models have adopted fully vectorized map representations, initially employing graph neural networks \cite{liang2020lanegcn, gao2020vectornet, ha2023learning, li2022graph} and Recurrent Neural Networks \cite{park2020diverse, chiara2022goal-driven, Lin2022Attention, ip2021vehicle}. The introduction of transformers in the field has further enhanced model capabilities, integrating attention mechanisms to refine trajectory predictions \cite{nayakanti_wayformer_2022, girgis_latent_2022, liu_laformer_2023, shi_motion_2023, Huang_2023_ICCV}.


Although there have been advances in model architectures aimed at enhancing prediction accuracy, the quality and safety of the diverse predicted trajectories have not been as thoroughly addressed. State-of-the-art models often produce undesirable results, such as trajectories that deviate off-road or violate traffic directions, as demonstrated in \cref{fig:pull}. Some models have incorporated explicit road structures to enhance scene feasibility \cite{deo_multimodal_2022, gilles_gohome_2021, gu_densetnt_2021}, and impose higher diversity \cite{Wang_2022_LTP, kim2023diverse, park2020diverse}. yet their dependency on specific map structures limits their adaptability across various datasets. Instead of developing new architectural frameworks, our work introduces simple universal loss functions that enhance prediction quality and diversity across any prediction model.

A significant body of work related to ours has also introduced an Offroad metric \cite{ridel2020scene, niedoba2019improving, boulton2020motion, messaoud2020multi, Argoverse, cui2021ellipse}, with some employing it directly as an auxiliary loss function \cite{niedoba2019improving, boulton2020motion, messaoud2020multi, cui2021ellipse}. These studies traditionally rely on rasterized offroad masks, which are less compatible with recent models. We address this by implementing our Offroad loss in a vectorized form, being more fine-grained and suitable for modern transformer-based models, and implement it using computational geometry algorithms on GPU for better performance.

\cite{greer2021trajectory} introduced a Yaw Loss that aligns vehicle headings with the nearest centerline, which proved suboptimal for complex scenarios like intersections involving multiple centerlines with varying directions. Their method had to define certain exceptions and excluded intersection cases completely. In contrast, our approach considers all potential centerline paths, applying penalties based on a blend of 2D positioning and heading distances, thus offering greater flexibility and applicability than prior methods.

Reinforcement learning-based methods, such as \cite{casas_importance_2020}, optimize motion prediction using reward signals to guide trajectory generation. However, these approaches suffer from high computational cost and instability during training due to the non-differentiable nature of their reward functions. In contrast, our proposed loss functions are fully differentiable, ensuring lower computational overhead and more stable training.

%% file: sections/method.tex
\begin{figure*}[!htbp]
    \centering
    \begin{subfigure}[b]{0.32\linewidth}
        \centering
        \includegraphics[width=\linewidth]{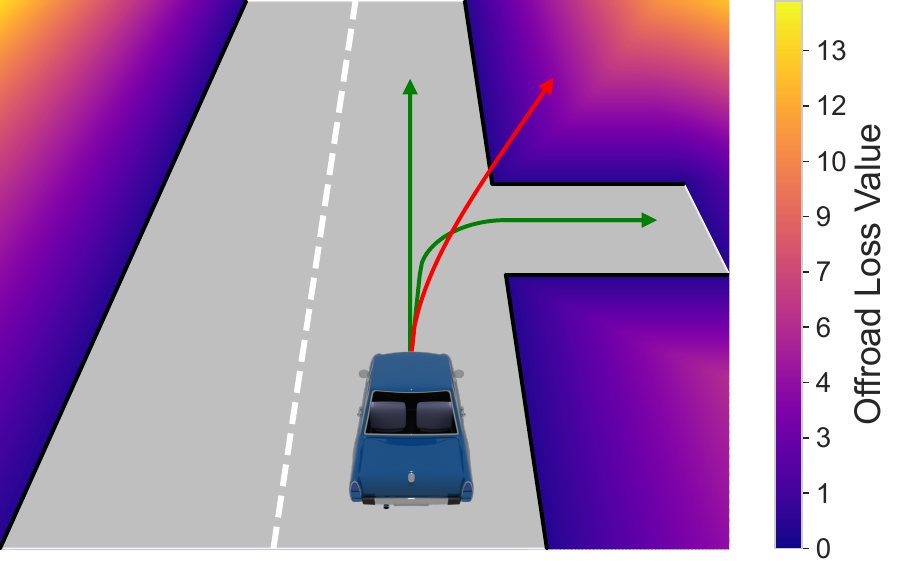}
        \caption{Offroad Loss}
        \label{fig:offroad_illustration}
    \end{subfigure}
    ~
    \begin{subfigure}[b]{0.32\linewidth}
        \centering
        \includegraphics[width=\linewidth]{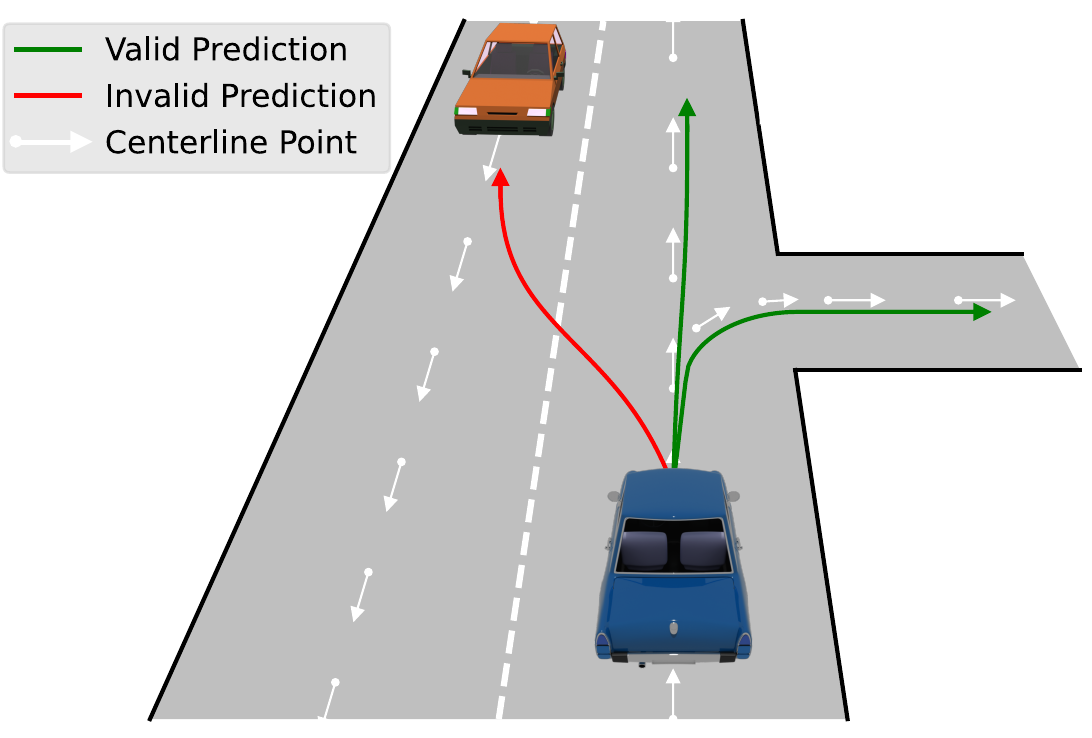}
        \caption{Road Direction Consistency}
        \label{fig:consistency_illustration}
    \end{subfigure}
    ~
    \begin{subfigure}[b]{0.32\linewidth}
        \centering
        \includegraphics[width=\linewidth]{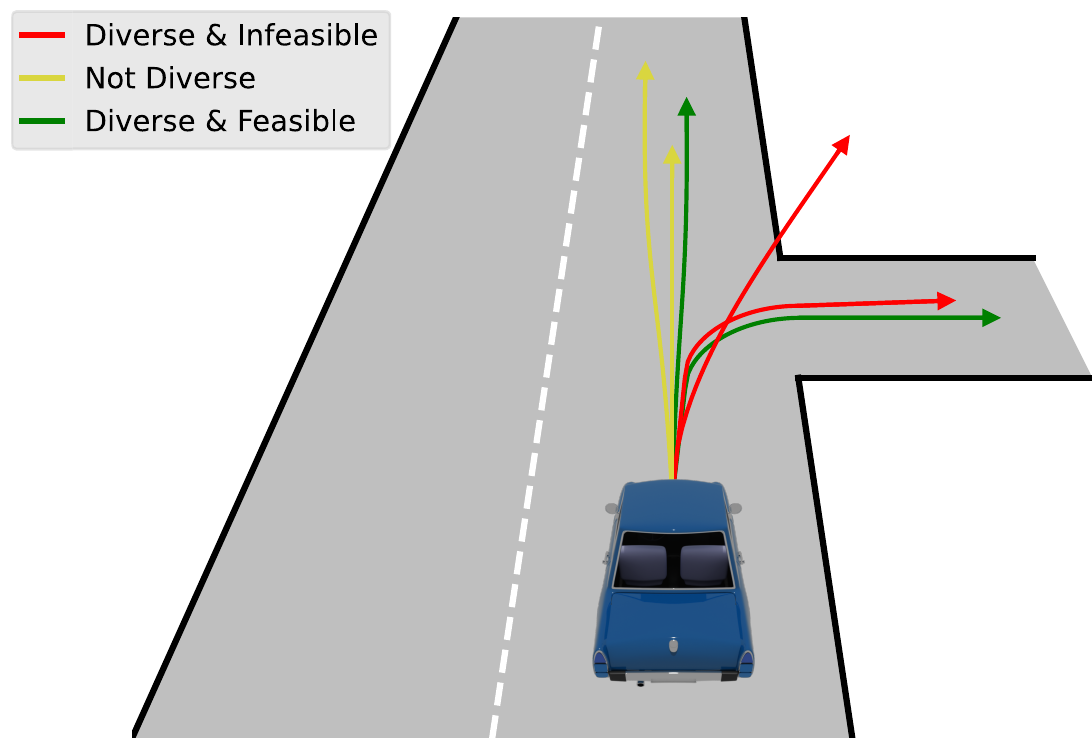}
        \caption{Mode Diversity}
        \label{fig:diversity_illustration}
    \end{subfigure}

    \caption{An illustration for our proposed loss functions. The colors in (a) show the Offroad Loss values for areas around the blue vehicle, with the red offroad trajectory having a high penalty. Panel (b) illustrates Road Direction Consistency, showing centerline points and directions; the incorrect red trajectory fails to align with the proper road direction. In (c), we compare three prediction sets: red trajectories demonstrate diversity but are infeasible due to straying off the drivable area; yellow trajectories are feasible but lack diversity, missing a potential right turn; green trajectories successfully combine diversity with feasibility, accurately reflecting viable path options. }
    \label{fig:loss_illustration}
    \vspace{-5pt}
\end{figure*}
This section formally defines the vehicle trajectory prediction problem to establish a mathematical framework. It then reviews the challenges inherent in the commonly used ``winner takes all" training approach in recent models. Finally, the section presents three novel auxiliary loss functions designed to enrich models with greater scene understanding and promote diversity in predictions. 

\subsection{Formulation}
Consider the trajectory prediction problem involving an ego agent surrounded by $N$ neighbouring agents within a scene. Let $s_t^i = (x_t^i, y_t^i)$ define the state of agent $i$ at timestep $t$. We have access to the observed trajectories $\bm x^i=[s_{-t}^i, \cdots, s_{-1}^i, s_0^i]$ for all agents over the past $t$ timesteps, aggregated as $\bm x = [\bm x^0, \bm x^1, \cdots, \bm x^N] \in \mathbb{R}^{N\times t \times 2}$. Additionally, the input includes the drivable area $\Omega$, represented as a union of several closed polygons, and the centerlines $\bm c \in \mathbb{R}^{S\times K \times 3}$, with each of $S$ centerline segments containing $K$ points characterized by their $2D$ position and yaw $(x, y, \theta)$. The prediction model $f$ aims to predict the ego agent's ground truth trajectory $\bm{\hat{y}} = [s_1^0, \cdots, s_T^0]$ over the next $T$ timesteps. Given the multimodal nature of the prediction task, stemming from uncertainties in drivers' intentions, the model $f$ generates $M$ distinct possible trajectories, thus $\bm y = f(\bm x, \Omega, \bm c) \in \mathbb{R}^{M\times T\times 2}$. 

\subsection{Previous objective functions}
Multimodal predictions are a fundamental aspect of current trajectory prediction models, with each model generating $M$ distinct trajectories. Traditionally, these models are trained using objective functions like minADE, defined as: 
\begin{align} \mathrm{minADE}=\min_{1\leq m \leq M} \frac{1}{T}\sum_{i=1}^T \norm{ \bm y_t^m - \bm{\hat{y}}_t^m}. \label{eq:minADE} \end{align} 
This function focuses on minimizing the error for the trajectory closest to the ground truth, effectively prioritizing the most likely outcome while allowing other predictions to represent alternative possible scenarios. However, such an approach has a significant drawback: only the closest trajectory to the ground truth receives gradient updates during training, leaving other predicted trajectories largely unrefined. This method of sparse supervision can lead to unsatisfactory modeling of less common but plausible behaviors, thereby limiting the model's ability to fully capture the diverse possibilities inherent in real-world driving scenarios.

\subsection{Proposed objective functions}
To address the limitations of common ``winner takes all" training objectives, we propose three new loss functions. Unlike traditional methods, these functions supervise all prediction modes, enhancing different aspects of the model's performance. Each function is designed to infuse the model with deeper knowledge, targeting specific shortcomings in existing approaches. \cref{fig:loss_illustration} illustrates these loss functions.

\subsubsection{\textbf{Offroad Loss}}
The first proposed loss function directs predictions toward the drivable area and penalizes off-road deviations by employing a signed distance function between the predicted trajectory $\bm y$ and the drivable area $\Omega$. This function is continuous and differentiable, with its gradient directed toward the nearest point within $\Omega$. \cref{fig:offroad_illustration} visually demonstrates the offroad function, where the penalty values are depicted across different areas around the vehicle.  Mathematically, the signed distance function $\phi$ for $\Omega$ is defined such that it returns the shortest distance from any point $x$ to the boundary $\partial \Omega$ of $\Omega$, being negative if $x$ is inside $\Omega$ and positive otherwise.  We define our Offroad loss function as:
\begin{align}
    \text{Offroad Loss}(\bm y, \Omega) = \frac{1}{M}\sum_{i=1}^M \sum_{t=1}^T \max(\phi(\bm y_t^i, \Omega) + m, 0),
\end{align}
where $\bm y_t^i$ represents the position at timestep $t$ of the $i$th prediction, $\phi$ is the signed distance function, and $m$ is a margin that maintains a buffer from $\Omega$'s boundary. This loss sums across all prediction points, with zero indicating presence within the drivable area (including a margin of $m$ meters) and increasing as predictions move further from $\Omega$.

To compute the signed distance function $\phi(\bm p, \Omega)$, we calculate the distance from point $\bm p$ to all polygon edges in $\Omega$ and select the minimum distance to determine the closest boundary. For determining whether $\bm p$ is inside $\Omega$, a ray casting algorithm is utilized, where a ray extending from $\bm p$ along the x-axis is used to count the intersections with the polygon's edges. Then, point $\bm p$ is in or out of $\Omega$ if the number of crossings is odd or even, respectively. This method, commonly used for point-in-polygon tests, is adapted from computational geometry principles outlined by \cite{orourke1998computational}. All computations are efficiently implemented on the GPU to minimize the performance overhead and accelerate the loss evaluation process.

\subsubsection{\textbf{Direction Consistency Error}}
Given that road centerlines are part of the map context for prediction models, maintaining the correct directional alignment is crucial. To address instances where models may align with these centerlines in the incorrect direction, we propose the Road Direction Consistency loss function. As \cref{fig:consistency_illustration} illustrates, this function imposes a significant penalty on trajectories deviating from corresponding centerline's direction, ensuring alignment not just in position but also in orientation.

To implement it, we calculate the heading direction $\gamma_t^i$ of each predicted trajectory $\bm y^i$ at timestep $t$. The difference $\delta(c, y_t^i)$ between a centerline point $c=(x, y, \theta)$ and a trajectory point $y_t^i=(x', y', \gamma_t^i)$ is defined as: 
\begin{align*}
    \delta(c, y_t^i) = &\max\big(\norm{(x, y) - (x', y')} - m_d, 0\big) \\+& \max\big(|\theta - \gamma_t^i| - m_\theta , 0\big),
\end{align*}
where $m_d \text{ and } m_\theta$ represent the allowable distance and angle margins, respectively.

The Road Direction Consistency loss then aggregates the minimum $\delta$ value between each trajectory point and all centerline points, summed across all trajectory points:
\begin{align}
    \text{Direction Consistency}(\bm y, \bm c) = \frac{1}{M}\sum_{i=1}^M \sum_{t=1}^T \min_{\substack{{1 \leq s \leq S}\\{1 \leq k \leq K}}} \delta(\bm c_k^s, \bm y_t^i).
\end{align}
The flexibility of this loss function is particularly important in complex environments like intersections, where many centerlines are close together but may have very different directions. Matching the trajectory strictly to the nearest centerline can sometimes lead to incorrect alignments. Our approach allows for matching with any centerline while introducing a penalty for the distance, ensuring a more accurate match. This means the model might align a trajectory with a centerline that is not the closest one but has a more suitable heading. This method ensures each trajectory point is evaluated against the most appropriate centerline, effectively improving accuracy in both position and direction, especially in challenging scenarios.

\subsubsection{\textbf{Mode Diversity}}
Ensuring a diverse set of predictions is critical for robust and safe route planning, particularly to capture all highly probable and plausible trajectories. To support this, we introduce the Mode Diversity loss, which promotes a spread of predictions and is depicted in \cref{fig:diversity_illustration}, that favours more spread predictions over the concentrated ones, as long as the predictions are feasible. This function first excludes any trajectories that are off-road by employing an indicator function $\mathds{1}(i)$ which assesses whether trajectory $\bm y^i$ remains within drivable area. The diversity loss is then calculated as the sum of pairwise distances between all remaining feasible trajectories:
\begin{align}
    \text{Mode Diversity}(y, \mathds{1}) = 
    \sum_{\substack{1 \leq i \leq M \\ \mathds{1}(i) = 1}} \text{ }
    \sum_{\substack{i \leq j \leq M \\ \mathds{1}(j) = 1}}
    \frac{1}{T}
    \sum_{t=1}^T
    \norm{\bm y_t^i - \bm y_t^j},
\end{align}
where $\mathds{1}(i)=1$ indicates that trajectory $\bm y^i$ is within drivable area. This selective approach ensures that the Mode Diversity loss only considers trajectories that are practical and safe, thereby avoiding an undesired increase in the loss value from trajectories pushed outside of road boundaries.

%% file: sections/experiment.tex
\begin{table*}[!htbp]
\centering
\vspace{4pt}
\caption{Performance of AutoBots and Wayformer on nuScenes and Argoverse 2 datasets with integration of our proposed loss functions alongside the original objective functions. The results show marked improvements in targeted metrics without compromising accuracy. Incorporating all three loss functions, our method shows a balanced improvement over all quality metrics compared to the baseline models.
}
\resizebox{\textwidth}{!}{%
    \begin{tabular}{l | c c c c c c | c c c c c c }
    \toprule
    & \multicolumn{6}{c|}{\textbf{nuScenes}} & \multicolumn{6}{c}{\textbf{Argoverse 2}}\\
            & minADE{$\downarrow$} & minFDE{$\downarrow$} & MR{$\downarrow$} & Offroad{$\downarrow$} & Direction{$\downarrow$} & Diversity{$\uparrow$} & minADE{$\downarrow$} & minFDE{$\downarrow$} & MR{$\downarrow$} & Offroad{$\downarrow$} & Direction{$\downarrow$} & Diversity{$\uparrow$}\\
    \midrule
Wayformer & 1.08 & 2.47 & 0.42 & 2.73 & 7.68 & 57 & 0.87 & 1.79 & 0.30 & 0.68 & 4.24 & 57\\
+ Offroad & 1.07 & 2.54 & 0.43 & \textbf{1.58} & 6.46 & 70 & 0.87 & 1.78 & 0.28 & \textbf{0.28} & 3.65 & 60\\
+ Direction & 1.13 & 2.60 & 0.43 & \textbf{1.56} & \textbf{5.27} & 57 & 0.87 & 1.80 & 0.28 & 0.34 & \textbf{1.89} & 58\\
+ Diversity & 1.10 & 2.55 & 0.46 & 2.83 & 8.22 & \textbf{74} & 0.88 & 1.81 & 0.31 & 0.69 & 4.29 & \textbf{64}\\
+ All & 1.13 & 2.61 & 0.43 & 1.67 & 5.47 & 67 & 0.89 & 1.85 & 0.30 & 0.44 & 2.54 & 63\\
\midrule
Autobots & 1.27 & 2.70 & 0.43 & 1.93 & 5.80 & 64 & 0.92 & 1.86 & 0.30 & 0.30 & 3.67 & 58\\
+ Offroad & 1.26 & 2.65 & 0.41 & \textbf{1.09} & 4.94 & 65 & 0.86 & 1.68 & 0.25 & \textbf{0.21} & 3.75 & 59\\
+ Direction & 1.26 & 2.64 & 0.41 & \textbf{1.04} & \textbf{4.21} & 67 & 0.88 & 1.73 & 0.26 & \textbf{0.20} & \textbf{1.70} & 60\\
+ Diversity & 1.27 & 2.69 & 0.49 & 1.81 & 5.96 & \textbf{93} & 0.87 & 1.72 & 0.27 & 0.31 & 3.79 & 62\\
+ All & 1.26 & 2.70 & 0.43 & \textbf{1.01} & \textbf{4.27} & 74 & 0.95 & 1.99 & 0.33 & \textbf{0.19} & 1.94 & \textbf{63}\\
    \bottomrule
    \end{tabular}
}
\label{tab:full_train}
\end{table*}
\begin{figure*}[!htbp]
  \resizebox{0.9\textwidth}{!}{%
  \includegraphics[width=0.86\textwidth]{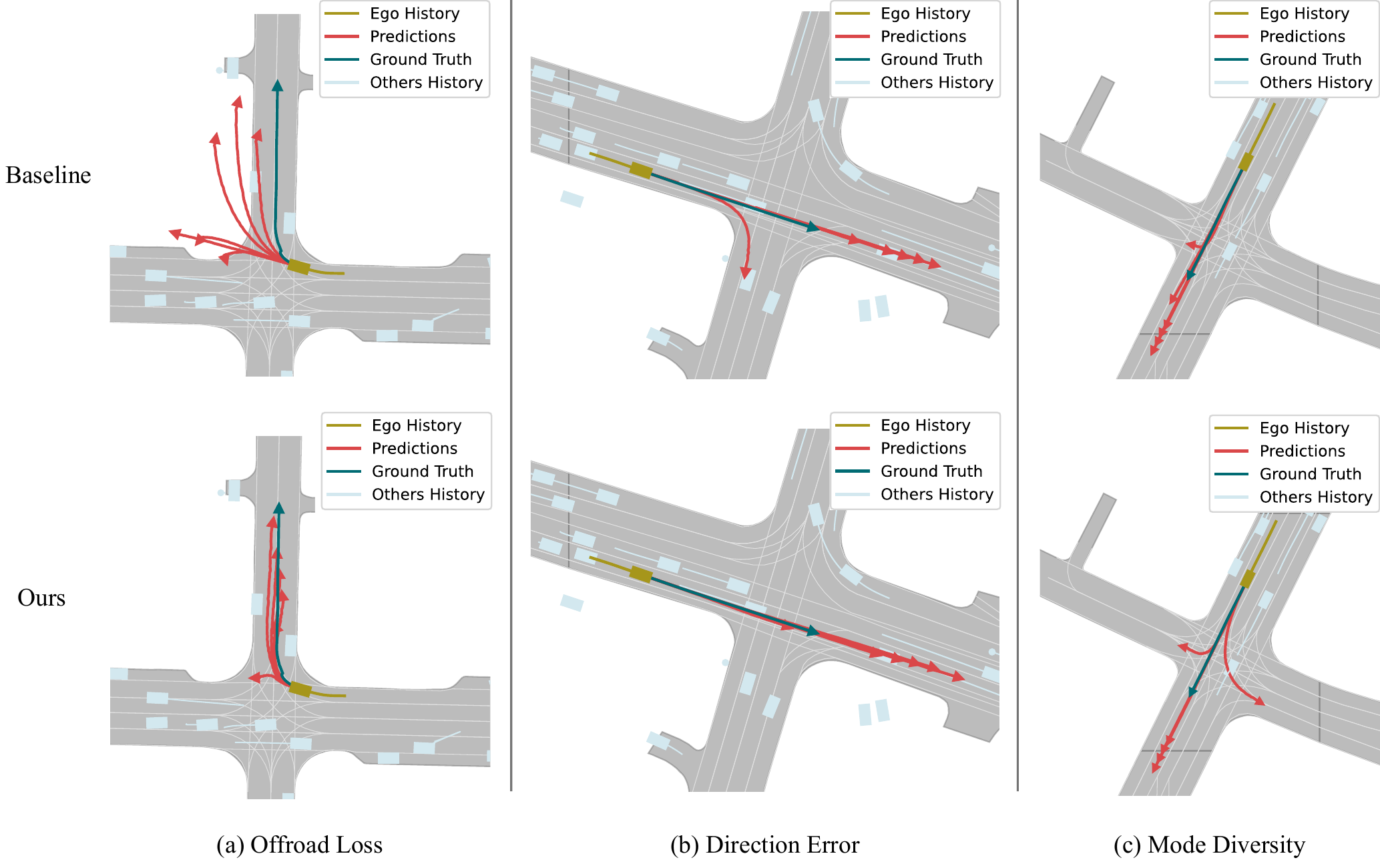}
  }
  \caption{
  Comparative visualization of model predictions on the Argoverse 2 dataset using Wayformer as the baseline. Each panel illustrates enhancements from applying a distinct auxiliary loss function to the baseline model: (a) Offroad Loss corrects four off-road predictions by enhancing adherence to drivable areas; (b) Direction Error adjusts a potentially hazardous right turn against traffic flow; (c) Mode Diversity introduces a new left turn and increases the spacing between trajectories. }
   \label{fig:qualitative}
   \vspace{-10pt}
\end{figure*}
\begin{figure*}[ht]
  \centering
  \begin{subfigure}[b]{0.3\linewidth}
        \centering
        \includegraphics[width=\linewidth]{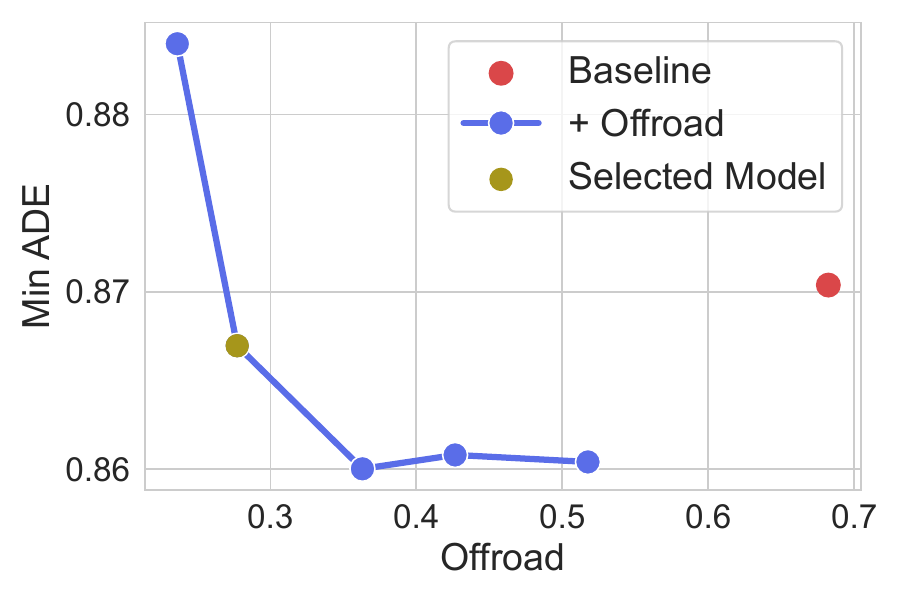}
        \caption{Offroad Loss $\alpha$ Curve}
        \label{fig:offroad_curve}
    \end{subfigure}
    ~
    \begin{subfigure}[b]{0.3\linewidth}
        \centering
        \includegraphics[width=\linewidth]{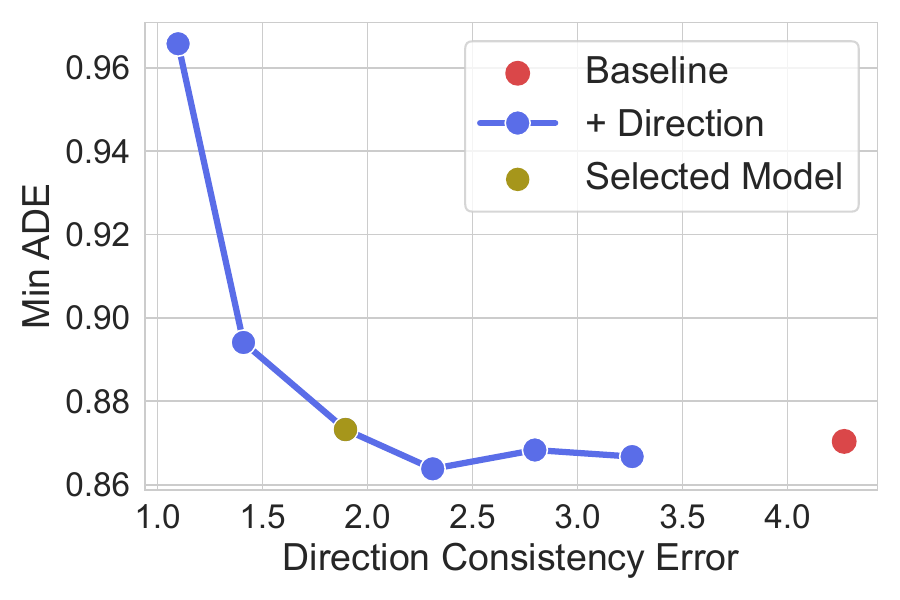}
        \caption{Direction Consistency $\alpha$ Curve}
        \label{fig:consistency_curve}
    \end{subfigure}
    ~
    \begin{subfigure}[b]{0.3\linewidth}
        \centering
        \includegraphics[width=\linewidth]{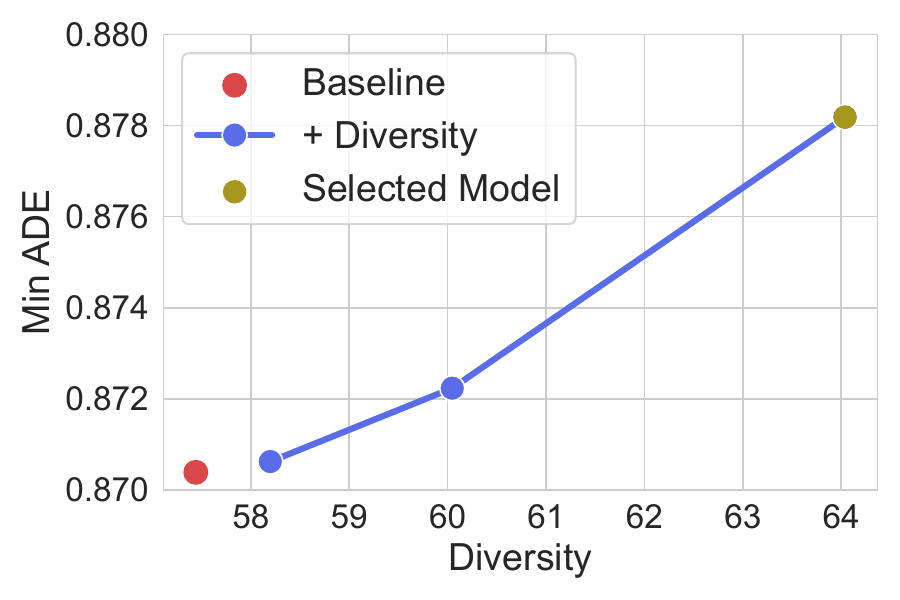}
        \caption{Diversity $\alpha$ Curve}
        \label{fig:diversity_curve}
    \end{subfigure}
    \caption{Performance impact of integrating our loss functions on Wayformer's minADE across the Argoverse 2 dataset. The blue curves demonstrate the trade-off between increasing auxiliary loss weights and prediction accuracy, with red dots marking the baseline performance. Yellow dots represent the selected models, showing regions where enhanced losses maintain similar accuracy compared to the baseline. Similar patterns are observed across different models and dataset configurations.
    }
    \label{fig:loss_curves}
    \vspace{-10pt}
\end{figure*}
In this section, we detail our experimental setup, including a description of the vehicle trajectory prediction datasets and the baseline models utilized. We enhanced the baseline models by integrating our proposed loss functions as auxiliary components. Specifically, the total loss for training is formulated as $\mathcal{L}_\mathrm{final} = \mathcal{L}_\mathrm{original} + \alpha \mathcal{L}_\mathrm{aux},$ where $\alpha$ is a hyperparameter that balances the contributions of the original and auxiliary losses. For experiments that utilize all introduced loss functions, we calculate the total auxiliary loss using a weighted sum of each function, where the weights are derived from the optimal $\alpha$ values determined during individual training with each specific loss function, and adjusted for optimal performance.  

We present both quantitative results demonstrating the performance improvements achieved with our auxiliary loss functions and qualitative examples that illustrate specific enhancements in trajectory prediction due to our methodology. Additionally, we explore how varying the weight $\alpha$ of the auxiliary loss affects overall model performance, providing insights into the optimal configuration for balancing between the baseline and proposed modifications. Finally, we show the improved robustness of our models through evaluations using the Scene Attack benchmark \cite{bahari2022sattack}, in realistic scenarios that include synthetically introduced turns.

\subsection{Experimental setup}
Our experiments leverage the UniTraj framework \cite{feng_unitraj_2024} to integrate our proposed loss functions with two state-of-the-art trajectory prediction models: Wayformer \cite{nayakanti_wayformer_2022} and AutoBots \cite{girgis_latent_2022}. Wayformer is known for its superior prediction capabilities within UniTraj, whereas AutoBots provides a performant yet lightweight alternative. The experiments are conducted on two prominent datasets: nuScenes \cite{caesar_nuscenes_2020}, a smaller and more challenging dataset, and Argoverse 2 \cite{chang_argoverse_2019}, which offers a broader range of scenarios. 

We adopt UniTraj's setup, where the models are trained using a history of 2 seconds and make predictions over 6 seconds, with each model outputting $M=6$ possible trajectories, using UniTraj's hyperparameters.

In terms of evaluation metrics, we introduce three novel measures — Offroad, Direction Error, and Diversity — to assess specific aspects of prediction quality. Additionally, we utilize standard metrics from the field which evaluate prediction accuracy, including minimum Average Displacement Error (minADE), as previously defined in \cref{eq:minADE}, along with minimum Final Displacement Error (minFDE) and Miss Rate (MR). MinFDE is calculated as:
\begin{align}
    \mathrm{minFDE}=\min_{1\leq m \leq M} \norm{ \bm y_T^m - \bm{\hat{y}}_T^m}.
\end{align}
Miss Rate (MR) is defined as the ratio of the samples where the minFDE exceeds 2 meters, and is useful where deviations up to 2 meters are acceptable.

\textbf{Training strategy:} We train the baseline models and those incorporating Offroad and Direction Consistency losses from scratch to achieve optimal results. However, the Diversity loss is applied through finetuning the baseline model over 10 epochs, as introducing it from the start tends to destabilize training. When employing all our loss functions together, we also opt for finetuning because of the Diversity loss’s impact, although this approach might slightly limit improvements in Offroad and Direction Consistency metrics.

\subsection{Quantitative results}
\label{sec:quantitative_results}
 The quantitative performance of our method, as detailed in \cref{tab:full_train}, demonstrates significant enhancements when training Wayformer and AutoBots on the nuScenes and Argoverse 2 datasets. We observe several key improvements: 
 \begin{itemize} 
    \item The implementation of our proposed loss functions consistently leads to substantial reductions in their targeted metrics. For instance, applying the Offroad loss typically cuts the offroad metric nearly in half. By integrating all three proposed loss functions, our method improves performance across all assessed quality metrics compared to baseline models.
    \item Although we observe a slight increase in minADE/ minFDE in certain settings, this is offset by the significant reduction in scene compliance errors like off-road, which is critical for ensuring safe and feasible predictions. This trade-off highlights the importance of prioritizing safety over minor decreases in accuracy in real-world autonomous driving scenarios.
    \item The Offroad and Direction Error metrics benefit each other. Using one as an auxiliary loss not only improves its own performance but also enhances the other. This mutual boost is especially notable with the Direction Error loss, which often shows improvements similar to those achieved with the Offroad loss.
\end{itemize}
\subsection{Qualitative results}

We present qualitative results of our proposed auxiliary loss functions in \cref{fig:qualitative}, demonstrating significant improvements in model predictions across various scenarios. In the first panel, the baseline model generates multiple off-road predictions, failing to align with the ground truth. Incorporating the Offroad loss during training teaches the model to recognize and avoid such errant paths, leading to more accurate trajectory predictions. In the second panel, a notable error in the baseline predictions includes a dangerous right turn against traffic flow. The application of our Direction Error loss corrects this, aligning the predictions with the correct traffic direction, thus enhancing safety and compliance with traffic rules. The third panel showcases the effects of our Mode Diversity loss, which significantly increases the spread of predicted trajectories. This loss function not only introduces additional plausible maneuvers, such as a new left turn that was absent in the baseline predictions but also ensures that the predicted trajectories are better spaced, reducing the likelihood of missing behaviors and increasing the overall prediction diversity.

\subsection{Metric weight study}
\label{sec:metric_weight_curve}

In this study, we explore how combining the original loss function with our auxiliary loss affects model performance. We illustrate this relationship in \cref{fig:loss_curves}, where we adjust the weight of the auxiliary loss, $\alpha$. Starting with a high auxiliary weight $\alpha$ close to the selected model in each curve, we gradually decrease this weight, moving towards baseline, which typically worsens the auxiliary metrics but can improve the main prediction accuracy.

Interestingly, there is a sweet spot along these curves where the prediction accuracy, specifically minADE, is close to that of the baseline model, while simultaneously improving on the auxiliary metrics. This balance demonstrates that it’s possible to enhance certain aspects of a model's predictions without degrading overall accuracy. These results show that advanced models are capable of achieving a desirable balance between accuracy and specific improvements, validating the effectiveness of our proposed loss functions as discussed in \cref{sec:quantitative_results}. 

\begin{table}[!thbp]
\centering
\caption{
Offroad metrics for baseline and our enhanced models under Scene Attack, demonstrating resilience against naturalistic scene perturbations.
}
    \begin{tabular}{l | c c | c c }
    \toprule
    & \multicolumn{2}{c|}{\textbf{nuScenes}} & \multicolumn{2}{c}{\textbf{Argoverse 2}}\\
            & Original & Attacked & Original & Attacked\\
    \midrule
Autobots & 1.93 & 4.69 & 0.30 & 1.19\\
+ Offroad & 1.09 & 3.34 & 0.21 & 1.01\\
+ All & \textbf{1.01} & \textbf{3.09} & \textbf{0.19} & \textbf{0.80}\\
\midrule
Wayformer & 2.73 & 8.89 & 0.68 & 4.16\\
+ Offroad & \textbf{1.58} & 7.41 & \textbf{0.28} & \textbf{1.69}\\
+ All & 1.67 & \textbf{6.93} & 0.44 & 3.53\\
\midrule
    \end{tabular}
\label{tab:robustness}
\vspace{-10pt}
\end{table}
\subsection{Robustness to scene attack}
We evaluate the enhanced robustness of our models using the Scene Attack benchmark \cite{bahari2022sattack}, which simulates naturalistic attacks by introducing varying turns in the road ahead of the ego agent. \cref{tab:robustness} displays the Offroad metric for predictions made by our baseline models on both original and manipulated scenes across the Argoverse 2 and nuScenes datasets. These metrics are averaged across different types and degrees of introduced road changes. Notably, models trained with our proposed loss functions consistently have lower Offroad metrics, showing superior resilience to these out-of-distribution scenarios and highlighting their robustness against realistic scene perturbations.

%% file: sections/conclusion.tex
This study has demonstrated the efficacy of integrating novel loss functions — Offroad Loss, Direction Consistency Error, and Diversity Loss — into trajectory prediction models to significantly enhance their quality and robustness across diverse driving scenarios. Our comprehensive evaluation on the nuScenes and Argoverse 2 datasets, utilizing state-of-the-art baseline models Wayformer and Autobots, has substantiated that these functions not only reduce undesirable behaviors such as off-road trajectories and direction violations but also improve the overall diversity and plausibility of predicted trajectories. The proposed loss functions stand out for their ability to be applied universally across different models and datasets without necessitating specialized architectural adaptations. As autonomous technologies continue to evolve, the methodologies developed in this work provide a scalable and efficient approach to improving the safety and reliability of trajectory predictions in complex urban environments.

A possible extension to this work would be the addition of other differential loss functions to this suite to further enhance the quality of predictions. Examples would be collision avoidance and kinematic feasibility of predictions. Another possibility for future investigation is the integration of a adaptive weight assigning algorithm to these loss functions, possibly based on model's state during training.